\documentclass[runningheads]{llncs}
\usepackage{graphicx}
\usepackage{amsmath,amssymb} 
\usepackage{epsfig}
\usepackage{subcaption}
\usepackage{multirow}
\usepackage{graphicx}
\usepackage{makecell}
 \usepackage{tikz}
\usepackage[misc]{ifsym} 

\usepackage{booktabs}
\usepackage{verbatim}  

\usepackage{color}

\usepackage[breaklinks=true,bookmarks=false]{hyperref}
\makeatletter
\newcommand{\printfnsymbol}[1]{%
  \textsuperscript{\@fnsymbol{#1}}%
}
\makeatother

\usepackage[T1]{fontenc}
\begin{document}
\title{Point Cloud Upsampling via Cascaded Refinement Network}
%
%
\author{Hang Du\thanks{Equal contribution.}
\and Xuejun Yan\printfnsymbol{1} 
\and Jingjing Wang 
\and Di Xie
\and Shiliang Pu$^{(\textrm{\Letter})}$}
\authorrunning{Du. et al.}
\titlerunning{Point Cloud Upsampling via Cascaded Refinement Network} 
%
\institute{Hikvision Research Institute, Hangzhou, China\\
\email{\{duhang, yanxuejun, wangjingjing9, xiedi, \\ pushiliang.hri\}@hikvision.com}}
\maketitle              

\begin{abstract}
Point cloud upsampling focuses on generating a dense, uniform and proximity-to-surface point set. 
Most previous approaches accomplish these objectives by carefully designing a single-stage network, which makes it still challenging to generate a high-fidelity point distribution. 
Instead, upsampling point cloud in a coarse-to-fine manner is a decent solution. 
However, existing coarse-to-fine upsampling methods require extra training strategies, which are complicated and time-consuming during the training. 
In this paper, we propose a simple yet effective cascaded refinement network, consisting of three generation stages that have the same network architecture but achieve different objectives.  
Specifically, the first two upsampling stages generate the dense but coarse points progressively, while the last refinement stage further adjust the coarse points to a better position. 
To mitigate the learning conflicts between multiple stages and decrease the difficulty of regressing new points, we encourage each stage to predict the point offsets with respect to the input shape.
In this manner, the proposed cascaded refinement network can be easily optimized without extra learning strategies. 
Moreover, we design a transformer-based feature extraction module to learn the informative global and local shape context. 
In inference phase, we can dynamically adjust the model efficiency and effectiveness, depending on the available computational resources. 
Extensive experiments on both synthetic and real-scanned datasets demonstrate that the proposed approach outperforms the existing state-of-the-art methods. 
The code is publicly available at \url{https://github.com/hikvision-research/3DVision}. 
\end{abstract}

\section{Introduction}
\label{sec:intro}
Point clouds have been widely adopted in many 3D computer vision studies~\cite{Qi2017PointNetDL,Qi2017PointNetDH,Yuan2018PCNPC,Li2018PointCNNCO,Wang2019DynamicGC,Zhao2020PointT,Long2021PC2PUPC} in recent years. However, in real-world scenarios, the raw point clouds produced by the 3D sensors are often sparse, noisy, and non-uniform, which have  negative impact on the performance of the point cloud analysis and processing tasks. 
Therefore, in order to facilitate the downstream point cloud tasks, it is necessary to upsample sparse point clouds to a dense, uniform and high-fidelity point set.

\begin{figure}[t]
    \centering
    \includegraphics[scale=0.28]{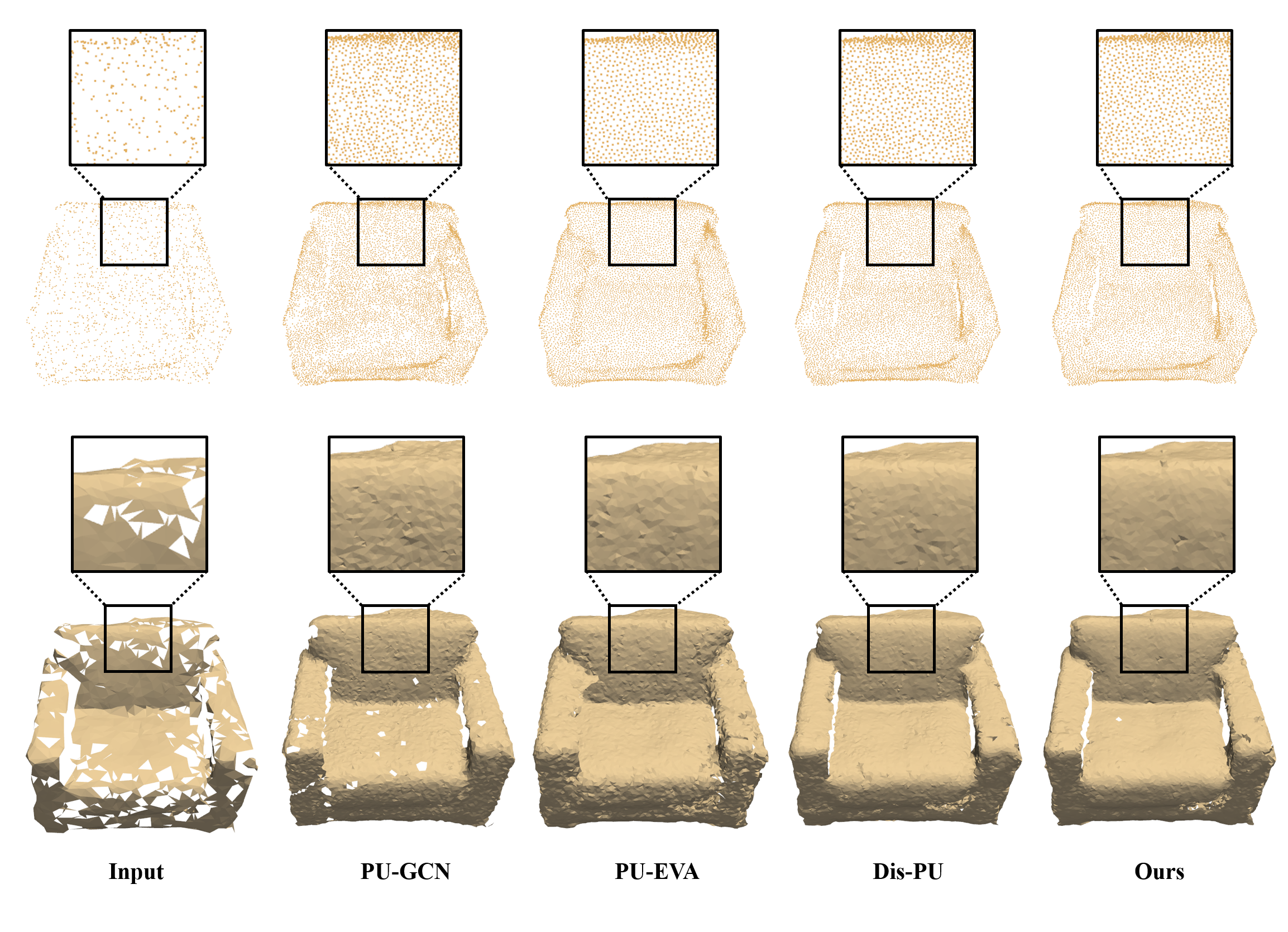}
    \caption{Upsampling results ($\times 4$) on real-scanned point clouds. The top line is the upsampled point clouds, and the bottom line shows the corresponding 3D surface reconstruction results using the ball-pivoting algorithm~\cite{bernardini1999ball}.  Compared with the recent state-of-the-art methods, we can generate 
    high-fidelity point clouds with uniform distribution, resulting in a more smooth reconstruction surface. 
    }
    \label{figure_1}
\end{figure}

In recent years, many deep learning-based methods have been proposed for point cloud upsampling. Compared with the traditional optimization-based methods~\cite{Alexa2003ComputingAR,Lipman2007ParameterizationfreePF,Huang2013EdgeawarePS,WuShihao2015DeepPC}, deep learning-based methods~\cite{Yu2018PUNetPC,Yu2018ECNetAE,Qian2020PUGeoNetAG,Li2019PUGANAP,Qian2021PUGCNPC} are able to handle more complex geometric structures of 3D objects since they can effectively extract deep features from point clouds and learn to generate new points in a data-driven manner. 
Among them, as a pioneering point cloud upsampling method, Yu~\textit{et al.}~\cite{Yu2018PUNetPC} propose a classic upsampling framework, named PU-Net, which develops a PointNet-based network to learn multi-scale point features and expand the number of points by multi-branch Multi-layer Perceptrons (MLP). 
Based on the framework of PU-Net, the recent advanced upsampling methods~\cite{Li2019PUGANAP,Qian2021PUGCNPC,LuoPUEVAAE} have made remarkable progress by designing more effective point feature extractor and upsampling unit. 
However, these methods directly generate the dense and high-fidelity points via a single-stage network, which makes the network challenging to meet multiple objectives and cannot obtain an optimal result through one-stage structure  (Fig.~\ref{figure_1}). 

Coarse-to-fine is a widely-used scheme in many point cloud generative models~\cite{yuan2018pcn,wang2020cascaded,Xie2020GRNetGR,xiang2021snowflakenet,Yan2022FBNet}. 
Certain methods~\cite{Wang2019PatchBasedP3,Li2021PointCU} also have applied such scheme for point cloud upsampling. Specifically,  MPU~\cite{Wang2019PatchBasedP3} contains a series of upsampling sub-networks that focus on different level of details. During the training, it needs to extract the local patches from the previous stage and activate multiple upsampling units progressively, which is complicated and time-consuming, especially on a large upsampling rate.
More recently, Li~\textit{et al.}~\cite{Li2021PointCU} propose Dis-PU to disentangle the multiple objectives of upsampling task, which consists of a dense generator for coarse generation and a spatial refiner for high-quality generation. 
In order to make the results more reliable, Dis-PU utilizes a warm-up training strategy that gradually increases the learning rate of the spatial refiner, which requires a longer training epoch to ensure full convergence of the networks. 

In the view of these limitations, we argue that a better coarse-to-fine point upsampling framework should be flexible and easily optimized during the training. 
To this end, we propose to accomplish coarse-to-fine point cloud upsampling via three cascaded generation stages, each of which has the same network architecture but focuses on different purposes. The overview of the proposed network is shown in Fig.~\ref{framework}. 
Among each generation stage, we leverage the local self-attention mechanism by designing a transformer-based feature extraction module that can learn both the global and local shape context.
As for feature expansion, we employ a two-branch based module to exploit the geometric information and learn shape variations rather than directly duplicating the initial point-wise features. 

In addition, the learning conflicts between multiple stages is a major challenge to the coarse-to-fine framework. 
To tackle this problem, we simply adopt a residual learning scheme for point coordinate reconstruction, which predicts the offset of each point and further adjusts the initial point position. In this manner, the entire cascaded refinement network enables to be optimized without extra training strategies. 

Resorting to the above practices, we can build a flexible and effective point cloud upsampling framework. 
Compared with the existing methods, the proposed framework can obtain significant performance improvements. 
According to the available computational resources, the feed-forward process of refinement stage is optional. Thus, we can dynamically adjust the model efficiency and effectiveness in inference phase. 
Extensive experiments on both synthetic and real-scanned datasets demonstrate the effectiveness of our proposed method.

The contributions of this paper are summarized as follows: 

\begin{itemize}
\item We propose a cascaded refinement network for point cloud upsampling. The proposed network can be easily optimized without extra training strategies. Besides, our method has better scalability to dynamically adjust the model efficiency and effectiveness in inference phase. 

\item We adopt the residual learning scheme in both point upsampling and refinement stages, and develop a transformer-based feature extraction module to learn both global and local shape context. 

\item We conduct comprehensive experiments on both synthetic and real-scanned point cloud datasets,  and achieve the leading performance among the state-of-the-art methods.

\end{itemize}

\section{Related Work}
Our work is related to the point cloud upsampling and the networks in point cloud processing. In this section, we provide a brief review of recent advances in these areas.

\subsection{Point Cloud Upsampling}
Existing point cloud upsampling methods can be roughly divided into traditional optimization-based approaches~\cite{Alexa2003ComputingAR,Lipman2007ParameterizationfreePF,Huang2013EdgeawarePS,WuShihao2015DeepPC} and the deep learning-based approaches~\cite{Yu2018PUNetPC,Yu2018ECNetAE,Wang2019PatchBasedP3,Li2019PUGANAP,Qian2020PUGeoNetAG,Ye2021MetaPUAA,Qian2021PUGCNPC,Zhao2021SSPUNetSP,LuoPUEVAAE,Qian2021DeepMU,Li2021PointCU}. The former generally requires the geometric prior information of point clouds (\textit{e.g.,} edges and normal), and achieves poor results on the complex shapes. In contrast, the deep learning-based approaches have been prevailing and dominated the recent state-of-the-arts. 
Particularly, PU-Net~\cite{Yu2018PUNetPC} proposes a classic learning-based upsampling framework that involves three components, including feature extraction, feature expansion (upsampling unit), and point coordinate reconstruction. Later on, many works~\cite{Li2019PUGANAP,Qian2020PUGeoNetAG,Qian2021PUGCNPC,Zhao2021SSPUNetSP,LuoPUEVAAE,Qian2021DeepMU} follow this classic framework and focus on improving it from many aspects, such as feature extractor, upsampling scheme, training supervision. For example, PU-GAN~\cite{Li2019PUGANAP} introduces a upsampling adversarial network with a uniform loss to ensure the point uniformity, and PU-GCN~\cite{Qian2021PUGCNPC} leverages graph convolutional networks for both feature extraction and expansion.
The above methods employ a single-stage network to accomplish the multiple objectives of point cloud upsampling. However, it is challenging to achieve all the objectives at the same time, and coarse-to-fine manner is a more appropriate solution. To this end, MPU~\cite{Wang2019PatchBasedP3} and Dis-PU~\cite{Li2021PointCU} propose to divide the upsampling process into multiple steps, however, they require progressive training or warm-up strategy to ensure a better optimization, which is complicated and time-consuming during the training.  
In this work, instead of using extra training strategies, we develop a more flexible and effective coarse-to-fine framework.

\subsection{Point Cloud Processing}
There are three major categories of point cloud processing networks, including projection-based, voxel-based, and point-based networks. 
Since the point clouds have no spatial order and regular structure, projection-based methods~\cite{tatarchenko2018tangent,Lang2019PointPillarsFE} and voxel-based methods~\cite{Xie2020GRNetGR,Zhou2018VoxelNetEL,Graham20183DSS} transform irregular point cloud to 2D pixel or 3D voxel, and then apply 2D/3D convolution to extract the regular representations. 
In contrast, point-based networks are designed to process the irregular point cloud directly, which is able to avoid the loss of shape context during the projection or voxelization. Among them, some methods~\cite{Qi2017PointNetDL,Qi2017PointNetDH} employ MLP-based structure to extract point-wise features, and others~\cite{Wang2019DynamicGC,Li2019DeepGCNsCG} utilize graph-based convolutional networks to aggregate the point neighbourhood information.
Besides, certain methods~\cite{Li2018PointCNNCO,Shi2019PointRCNN3O} develop continuous convolutions that can be directly applied to the point cloud. 
More recently, the success of vision transformer has inspired many 3D point cloud processing approaches~\cite{Guo2021PCTPC,Engel2021PointT,Pan20213DOD,Zhao2020PointT,yu2021pointr}. 
As far as we can see, the transformer-based structure is under-explored in point cloud upsampling. Thus, based on the architecture of point transformer~\cite{Zhao2020PointT}, we build a transformer-based feature extraction module in our  upsampling network.

\section{Proposed Method}
\begin{figure*}[t]
    \centering
    \includegraphics[scale=0.36]{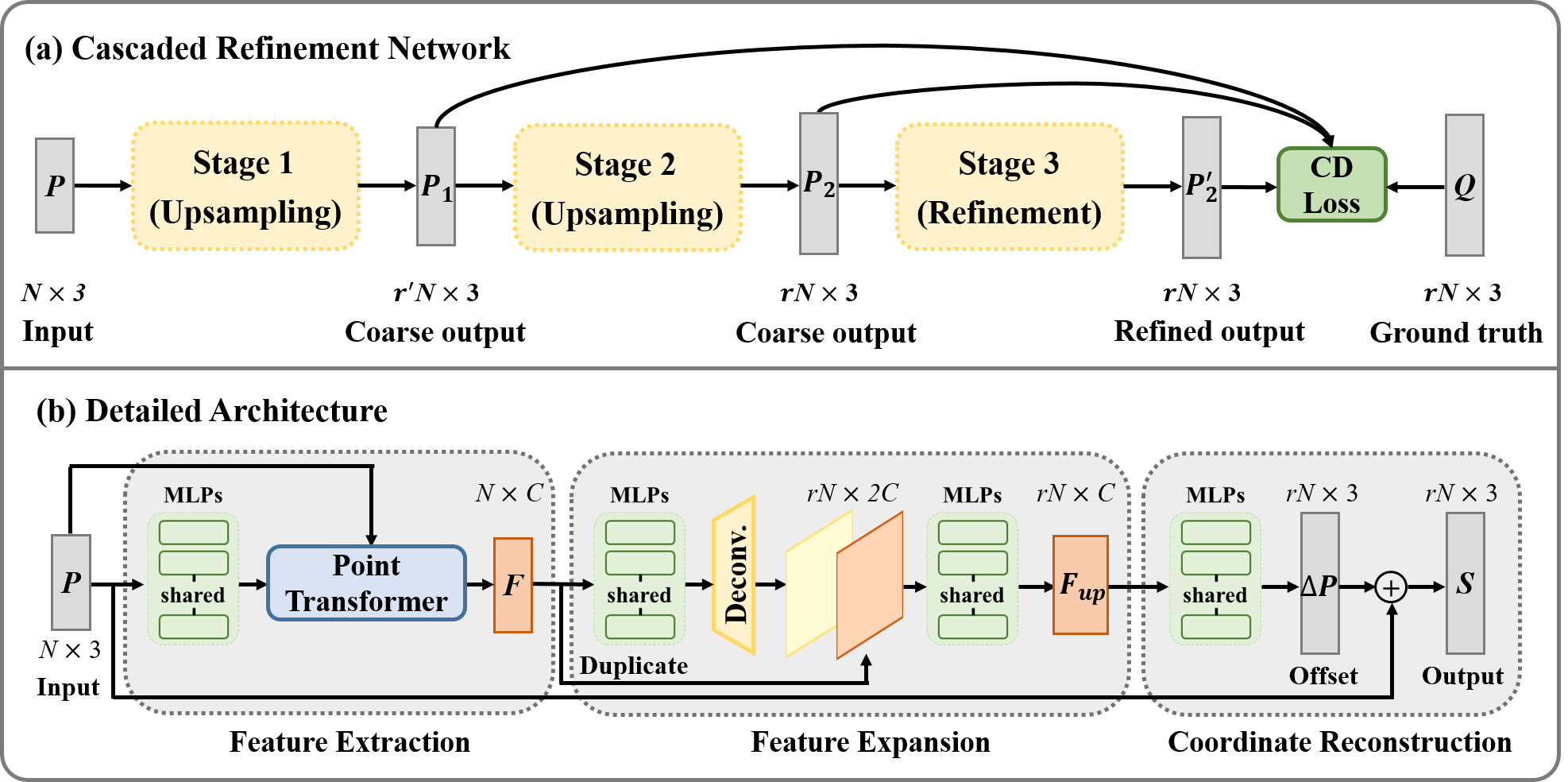}
    \caption{
    (a) Overview of our cascaded refinement network for point cloud upsampling, which consists of two upsampling stages and one refinement stage. 
    (b) For each upsampling or refinement stage, the detailed network architecture contains a transformer-based feature extraction module, a two-branch feature expansion module, and a coordinate reconstruction module.}
    \label{framework}
\end{figure*}

Given a sparse point cloud set $\mathcal{P}=\left\{p_{i}\right\}_{i=1}^{N}$, where $N$ is the number of points and $p_{i}$ is the 3D coordinates, the objective of point clouds upsampling is to generate a dense and high-fidelity point cloud set $\mathcal{S}=\left\{s_{i}\right\}_{i=1}^{rN}$, where $r$ is the upsampling rate.  
Fig.~\ref{framework} shows the overview of the proposed cascaded refinement network, where the first two upsampling stages generate a coarse dense point cloud set progressively and the last refinement stage servers as a refiner to adjust the coarse points to a better position.
During the training, we train the entire framework in an end-to-end manner, and adopt three Chamfer Distance (CD) losses to constrain each stage,  simultaneously. 
In the following sections, we elaborate the detailed network architecture of our framework and the training loss function.

\subsection{Network Architecture}
As shown in the bottom of Fig.~\ref{framework}, the network architecture of upsampling or refinement stage follows the commonly-used pipeline, which consists of a feature extraction module, a feature expansion module and a coordinate reconstruction module. 
In the following, we will provide a detailed introduction of each module, respectively.

\begin{figure*}[t]
    \centering
    \includegraphics[scale=0.42]{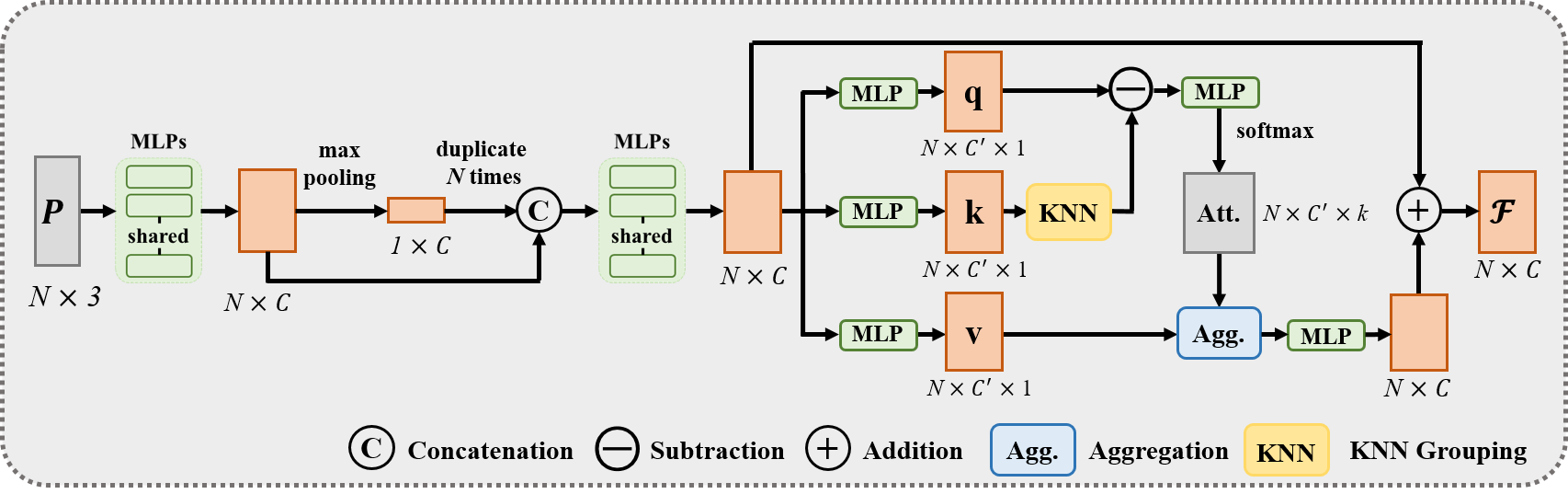}
    \caption{The detailed network architecture of transform-based feature extraction module. The attention map is calculated locally within the k-nearest neighbors of the current query. We omit the position encoding in value vectors and attention map for the clearness. 
    }
    \label{figure_feature}
\end{figure*}

\noindent \textbf{Feature extraction.} 
Given a input point cloud $\mathcal{P}=\left\{p_{i}\right\}_{i=1}^{N}$, the feature extraction module aims to encode point features $\mathcal{F}=\left\{f_{i}\right\}_{i=1}^{N}$ of $C$ channel dimensions. 
Most existing point cloud upsampling methods~\cite{Yu2018PUNetPC,Li2019PUGANAP,Li2021PointCU} fulfill this objective using a MLP-based structure. 
However, the local details are limited by the insufficient feature representation capability of MLPs. 
To solve this problem, some upsampling approaches~\cite{Wang2019DynamicGC,Li2019DeepGCNsCG,Qian2021PUGCNPC} employ a GCN-based structure to capture the local neighborhood information for point feature extraction.

In the view of the success of transformer-based networks~\cite{Pan20213DOD,Zhao2020PointT,Guo2021PCTPC} in the field of 3D computer vision, we explore a transformer-based feature extraction module for the task of point cloud upsampling. 
Fig.~\ref{figure_feature} illustrates the details on our proposed network architecture.
Based on the network architecture in~\cite{Qi2017PointNetDH}, we first extract the point-wise features from the input point clouds via a set of MLPs, and then apply a max pooling operation to obtain the global features. 
The global features are duplicated $N$ times and concatenated with the initial features, subsequently. 
After that, a point transformer layer~\cite{Zhao2020PointT} is utilized to refine the local shape context and obtain the output point features. 
By applying self-attention locally, the learned features can incorporate the point neighborhood information between the point feature vectors. Finally, the encoded point features $ \mathcal{F} =\left\{\mathbf{f}_{i}\right\}_{i=1}^{N}, \mathbf{f}_{i} \in \mathbb{R}^{C}$ are added with the input features through a residual connection.   
The locally self-attention operations can be formulated as: 

\begin{equation}
\label{attention}
\mathbf{f}_{i} = \mathbf{x}_{i} \oplus \sum_{{j} \in \mathcal{N}(i) } {softmax}\left(\mathbf{q}_{i}-\mathbf{k}_{j}\right) \odot\mathbf{v}_{j},
\end{equation}
where $\mathbf{x}_{i}$ is ${i}$th input point feature vector, $\mathbf{q}_{i}$ is the corresponding query vector, $\mathbf{k}_{j}$ is the ${j}$th key vector from the k-nearest neighbors of $\mathbf{x}_{i}$, and $\mathbf{v}_{j}$ is the ${j}$th value vector. The attention weights are calculated between the query and its k-nearest neighbors by softmax function.
Note that we omit the position encoding in Equation~\ref{attention}. One can refer to~\cite{Zhao2020PointT} for more details of this part. In this way, we are able to encode the point features by combining the local and global information.

\noindent \textbf{Feature expansion.} Subsequently, the extracted point features are feed into a feature expansion module that produces the expanded point feature $\mathcal{F_\text{up}}=\left\{f_{i}\right\}_{i=1}^{rN}$, where $r$ is the upsampling rate. 
Most previous upsampling approaches apply duplicate-based~\cite{Wang2019PatchBasedP3,Li2019PUGANAP,Li2021PointCU} or shuffle~\cite{Qian2021PUGCNPC} operation to increase the numbers of the input points features. 
In this work, we employ a two-branch based scheme for feature expansion, which combines the advantages of duplicate-based~\cite{Wang2019PatchBasedP3,Li2019PUGANAP} and learning-based methods~\cite{Hui2020ProgressivePC,Xie2020GRNetGR,xiang2021snowflakenet}. 
Specifically, we first duplicate the input features $\mathcal{F}=\left\{f_{i}\right\}_{i=1}^{N}$ with $r$ copies. 
Meanwhile, a transposed convolution branch is used for feature expansion, which is utilized to learn new point features through a learnable manner. 
Certain methods~\cite{Hui2020ProgressivePC,xiang2021snowflakenet} have discussed the advantages of learning-based feature interpolation for generative models, which can exploit more local  geometric structure information with respect to  the input shape. 
Finally, we concatenate the point features from two branches and feed them to a set of MLPs which produces the expanded features as follows: 

\begin{equation}
     \mathcal{F_\text{up}} = \text{MLP}( \text{Concat}[\text{Dup}(\mathcal{F}, r) ;\text{Deconv}(\mathcal{F})] ).
\end{equation}

The duplicated-based branch is able to preserve the initial shape, and the learnable transposed convolution branch can provide local geometric variations on the input point clouds. The combination of them enables to produce a more expressive upsampled features for reconstructing the point coordinate reconstruction subsequently.

\noindent \textbf{Coordinate reconstruction.}
The objective of coordinate reconstruction is to generate a new point set  $\mathcal{S}=\left\{s_{i}\right\}_{i=1}^{rN}$ from the expanded feature vectors $\mathcal{F_\text{up}}=\left\{f_{i}\right\}_{i=1}^{rN}$. 
To accomplish this objective, a common way is to regress the 3D point coordinates directly. 
However, it is difficult to generate high-fidelity points from latent space without noises~\cite{Li2021PointCU}. 
In order to solve this problem, several methods~\cite{Wang2019PatchBasedP3,Li2021PointCU} apply a residual learning strategy that predicts the offset $\Delta \mathcal{P}$ of each point to adjust the initial position, which can be formulated as: 
\begin{equation}
     \mathcal{P}_\text{output} = \Delta \mathcal{P} + \text{Dup}(\mathcal{P}_\text{input}, r), 
\end{equation}
where $r$ is the upsampling rate. 
Moreover, residual learning is a simple yet effective scheme to mitigate the learning conflicts between multiple stages in a coarse-to-fine framework. 
Thus, we consider the advantages of such scheme are two folds: 
(i) learning per-point offset can greatly decrease the learning difficulty of a cascaded network and mitigate the conflicts between multiple stages, which results in a stable and better optimization;  
(ii) residual connection enables to preserve the geometric information of initial shape and provide variations for generating new points. 
Hence, we adopt residual learning scheme for coordinate reconstruction, which contains two MLP layers to generate the per-point offset $\Delta \mathcal{P}$ from $\mathcal{F_\text{up}}$ ($\Delta \mathcal{P} = \text{MLP}(\mathcal{F_\text{up}}) $), and then adds it on the $r$ times duplicated input points. In this manner, the entire cascaded refinement network can be easily optimized together without extra training strategies (\textit{e.g.,} progressive or warm-up training).

\subsection{Compared with Previous Coarse-to-Fine Approaches}
\label{discussion}

The coarse-to-fine scheme has been studied in the previous point cloud upsampling methods~\cite{Wang2019PatchBasedP3,Li2021PointCU}. 
Nonetheless, the proposed approach is more efficient and flexible during the training and inference. 
Specifically, MPU~\cite{Wang2019PatchBasedP3} needs to extract the local training patches from the previous stage, and activates the training of each upsampling stage progressively. 
Dis-PU~\cite{Li2021PointCU} adopts a warm-up training strategy to control the importance of dense generator and spatial refiner in different training stage. 
Both of them are complex during the training and cost a longer time to ensure the full convergence of the networks.
Moreover, the complicated training strategies rely on the experience of hyper-parameters tuning, which may be sensitive on the new datasets.  
In contrast, our advantages are two folds. 
Firstly, the proposed cascaded refinement network can be easily trained without progressive or warm-up training strategies, which enables to require less training consumption. 
Secondly, our framework is more flexible and scalable.
We can dynamically adjust the model efficiency and effectiveness at the stage of inference. 
For example, we can pursue a faster inference time only using two-stage upsampling structure, and still achieve a leading performance among other competitors (Table~\ref{Complexity}).  
Therefore, the proposed coarse-to-fine upsampling framework can be more effective and practical in real-world applications.

\subsection{Training Loss Function}
During the training, our approach is optimized in an end-to-end manner. To encourage the upsampled points more distributed over the target surface, we employ Chamfer Distance (CD)~\cite{fan2017point} 
$\mathcal{L}_{CD}$ for training loss function, which is formulated as: 

\begin{equation}
    \mathcal{L}_{CD}\left(\mathcal{P}, \mathcal{Q}\right) = \frac{1}{|\mathcal{P}|} \sum_{p \in \mathcal{P}} \min _{q \in \mathcal{Q}} \left\|p-q\right\|^{2}_{2} + \frac{1}{|\mathcal{Q}|} \sum_{q \in\mathcal{Q}} \min _{p \in \mathcal{P}} \left\|p-q\right\|^{2}_{2},
\end{equation}
where $\mathcal{P}$ and $\mathcal{Q}$ denote the generated point clouds and their corresponding ground truth, respectively.  
CD loss aims to calculate the average closest point distance between the two sets of point cloud. 

In order to explicitly control the generation process of point clouds, we employ the CD loss on three generation stages, simultaneously. Therefore, the total training loss function $\mathcal{L}_{\text{total}}$ can be defined as: 
\begin{equation}
    \mathcal{L}_{\text{total}}=\mathcal{L}_{CD}\left(\mathcal{P}_{1}, \mathcal{Q}\right)+ \mathcal{L}_{CD} \left(\mathcal{P}_{2}, \mathcal{Q}\right) + \mathcal{L}_{CD} (\mathcal{P}^{'}_{2}, \mathcal{Q}), 
\end{equation}
where $\mathcal{P}_{1}$, $\mathcal{P}_{2}$ and $\mathcal{P}^{'}_{2}$ are the output point clouds from three generation stages, respectively. We validate the effectiveness of the penalty on three stages in the experiment (please refer to supplementary materials). The results show that the simultaneous supervision on three generation stages results in a better performance. 
Therefore, we adopt CD loss on three generation stages simultaneously.

\section{Experiments}

In this section, we conduct extensive experiments to verify the effectiveness of the proposed point upsampling network. 
The section is organized as follows. Section~\ref{setting} introduces the detail experimental settings. Section~\ref{results_general} demonstrates the obvious improvements by our approach on two synthetic datasets. Section~\ref{results_real} shows the qualitative comparisons on real-scanned data. Section~\ref{results_model} provides model complexity analysis with other counterparts.  Section~\ref{results_study} studies the effect of major components in our approach.

\subsection{Experimental Settings}
\label{setting}

\noindent \textbf{Datasets.}
We employ two point cloud upsampling datasets, including PU-GAN~\cite{Li2019PUGANAP} and PU1K~\cite{Qian2021PUGCNPC}.  
Among them, PU-GAN dataset contains 24,000 training patches collected from 120 3D models and 27 3D models for testing. 
In contrast, PU1K dataset is a large-scale point cloud upsampling dataset which is newly released by PU-GCN~\cite{Qian2021PUGCNPC}. PU1K dataset consists of 69,000 training patches collected from 1,020 3D models and 127 3D models for testing. 
In addition, we also utilize a real-scanned point cloud dataset ScanObjectNN~\cite{Uy2019RevisitingPC} for qualitative evaluation.

\noindent \textbf{Training and evaluation.} 
Our models are trained with 100 epochs. The batch size is set as 64 for PU1K and 32 for PU-GAN, respectively. The learning rate begins at 0.001 and drops by a decay rate of 0.7 every 50k iterations. 
The ground truth of each training patch contains 1,024 points and the input contains 256 points that are randomly sampled from the ground truth. 
In practical implementation, we set the upsampling rate $r=2$ in the upsampling stage and $r=1$ in the refinement stage, to achieve $\times 4$ upsampling. 
More detailed settings of our network can be found in the supplementary material.  
For inference, we follow the common settings~\cite{Li2019PUGANAP,Qian2021PUGCNPC} to divide the input point clouds into multiple patches based on the seed points. Then, the patches are upsampled with $r$ times. After that, FPS algorithm is used to combine all the upsampled patches as the output point clouds.
For quantitative evaluation, we apply three commonly used metrics, including Chamfer Distance (CD), Hausdorff Distance (HD), and Point-to-Surface Distance (P2F). A smaller value of these metrics denotes a better performance.

\noindent \textbf{Comparison methods.}
In the following, we will compare the proposed method with five existing point cloud upsampling methods, including PU-Net~\cite{Yu2018PUNetPC}, MP-U~\cite{Wang2019PatchBasedP3}, PU-GAN~\cite{Li2019PUGANAP}, Dis-PU~\cite{Li2021PointCU},  PU-GCN~\cite{Qian2021PUGCNPC}, and PU-EVA~\cite{LuoPUEVAAE}. 
For a fair comparison, we reproduce these methods by their officially released codes and recommended settings on the same experimental environment.

\subsection{Results on Synthetic Dataset}
\label{results_general}

\begin{table}[t]
    \begin{center}
    \caption{Quantitative comparison ($\times$4 upsampling) on PU1K dataset with different input sizes of point cloud. The values of CD, HD, and P2F are multiplied by $10^{3}$. A smaller value denotes a better performance. }
    \label{PU1K_performance}
    \resizebox{1\linewidth}{!}{
    \begin{tabular}{|l||p{1cm}<{\centering}p{1cm}<{\centering}p{1cm}<{\centering}|p{1cm}<{\centering}p{1cm}<{\centering}p{1cm}<{\centering}|p{1cm}<{\centering}p{1cm}<{\centering}p{1cm}<{\centering}|}   
        \hline
   \multirow{2}{*}{Methods}
   & \multicolumn{3}{c|}{{Sparse (512) input} }  
   & \multicolumn{3}{c|}{{Medium (1,024) input}} 
   & \multicolumn{3}{c|}{{Dense (2,048) input} }  \\ 
   & \makecell[c] {{CD}}
   & \makecell[c] {{HD}} 
   & \makecell[c] {{P2F}}
   & \makecell[c] {{CD}}
   & \makecell[c] {{HD}} 
   & \makecell[c] {{P2F}}
   &  \makecell[c] {{CD}}
   & \makecell[c] {{HD}} 
   & \makecell[c] {{P2F}}
       \\
    \hline\hline
    PU-Net~\cite{Yu2018PUNetPC} &2.990 &35.403 &11.272 &1.920&24.181&7.468 &1.157 & 15.297 & 4.924 \\
    MPU~\cite{Wang2019PatchBasedP3}  &2.727 & 30.471& 8.190 &1.268&16.088&4.777  &0.861 & 11.799 &3.181 \\
    PU-GAN~\cite{Li2019PUGANAP}  &2.089 &22.716 & 6.957 &1.151&14.781& 4.490 &0.661 &9.238 & 2.892 \\
    Dis-PU~\cite{Li2021PointCU} &2.130&25.471&7.299  &1.210&16.518&4.606 &0.731&9.505&2.719  \\
    PU-GCN~\cite{Qian2021PUGCNPC} & 1.975&22.527 & 6.338 &1.142&14.565&4.082  &0.635 &9.152 & 2.590 \\
    PU-EVA~\cite{LuoPUEVAAE} &1.942&20.980& 6.366&1.065&13.376& 4.172&0.649&8.870&  2.715\\ 
   \hline  \hline
    Ours &\textbf{1.594} &\textbf{17.733} & \textbf{4.892}   &\textbf{0.808}&\textbf{10.750}& \textbf{3.061}& \textbf{0.471} &\textbf{7.123}&\textbf{ 1.925 } \\
    \hline
    \end{tabular}}
    \end{center}

\end{table}

\begin{figure*}[t]
    \centering
    \includegraphics[scale=0.33]{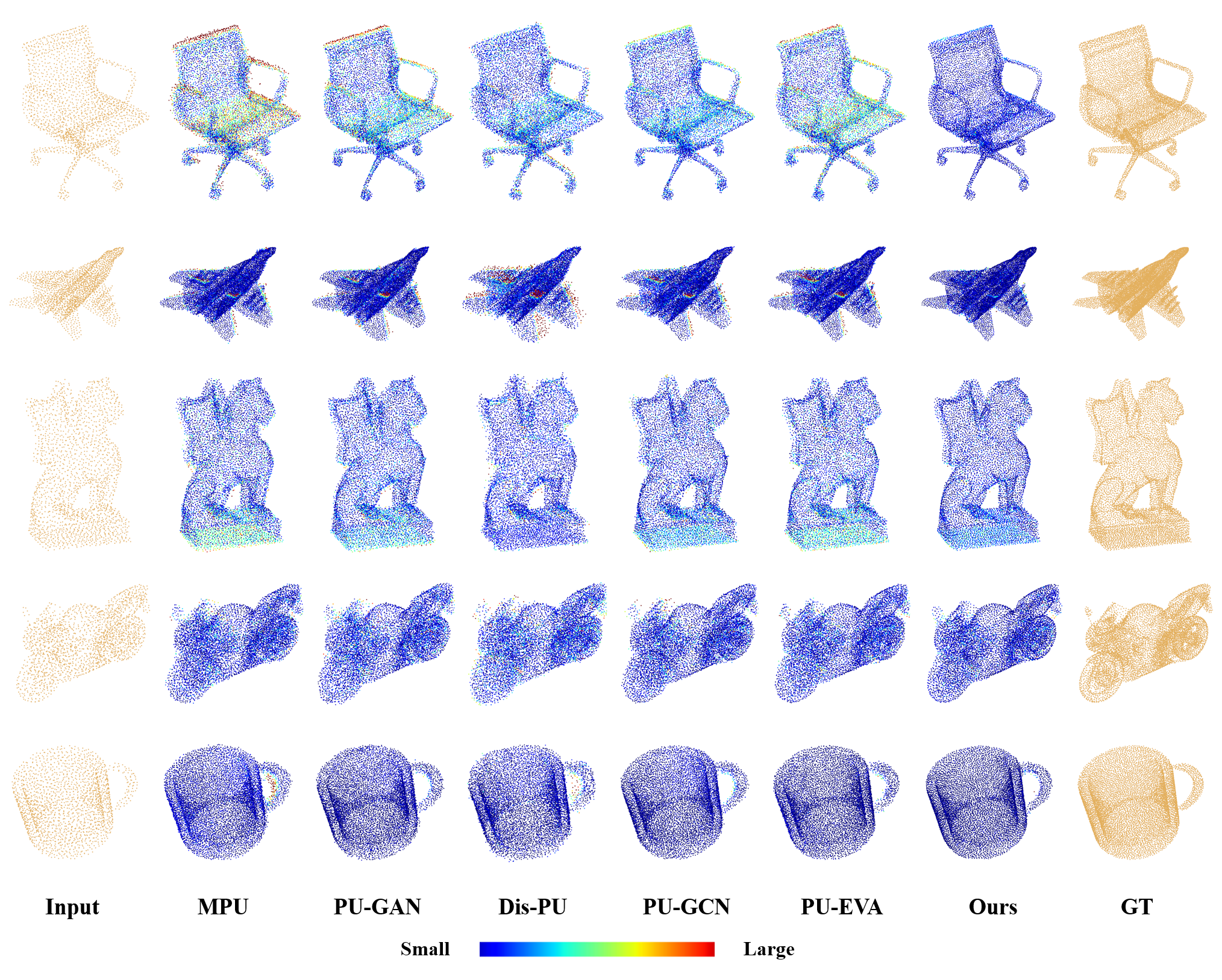}
    \caption{Point cloud upsampling ($\times 4$) results on PU1K dataset. The point clouds are colored by the nearest distance between the ground truth and the prediction of each method. The blue denotes the small errors. One can zoom in for more details. }
    \label{puk1_vis}
    
\end{figure*}

\noindent \textbf{PU1K dataset.} 
As shown in Table~\ref{PU1K_performance}, we conduct experiments on three different input sizes of point, including sparse (512), medium (1,024), and dense (2,048). 
From the results, we can observe the proposed method can achieve the leading performance in all metrics and obvious improvements on the sparse input. 
Besides, compared the results of left three columns (sparse input) with the right three columns (dense input), we can observe that the sparse input size is a more challenging scenario. 
Nonetheless, our method can consistently yield evident improvements over the existing state-of-the-art methods, which verifies the robustness of our method to different input sizes. 
Moreover, the qualitative results in Fig.~\ref{puk1_vis} can also demonstrate that our cascaded refinement network enables to generate more uniform point clouds with better fine-grained details and less outliers, such as the arms of the chair and the wings of the plane. 

\begin{table}[t]
    \begin{center}
    \caption{Quantitative comparison on PU-GAN~\cite{Li2019PUGANAP} dataset. The input size of point cloud is 1,024. The values of CD, HD, and P2F are multiplied by $10^{3}$. }
    \label{PU_GAN_performance}
    \resizebox{0.675\linewidth}{!}{
    \begin{tabular}{|l||ccc|ccc|}
    \hline
   \multirow{2}{*}{Methods}
   & \multicolumn{3}{c|}{{$\times$4 Upsampling} } 
   & \multicolumn{3}{c|}{{$\times$16 Upsampling} }  \\ 
   & \makecell[c] {{CD}}
   & \makecell[c] { {HD}} 
   & \makecell[c] {{P2F}}
   &  \makecell[c] {{CD}}
   & \makecell[c] { {HD}} 
   & \makecell[c] {{P2F}}
       \\
    \hline\hline
    PU-Net~\cite{Yu2018PUNetPC} &0.883 &7.132&8.127 &0.845& 11.225& 10.114 \\
    MPU~\cite{Wang2019PatchBasedP3}  &0.589 &6.206 &  4.568&0.365&8.113 &5.181 \\
    PU-GAN~\cite{Li2019PUGANAP}  & 0.566&6.932&4.281&0.390 &8.920 &4.988  \\
    Dis-PU~\cite{Li2021PointCU} & 0.527& 5.706& 3.378&0.302 &6.939 &4.146 \\
    PU-GCN~\cite{Qian2021PUGCNPC} & 0.584 &\textbf{5.257}& 3.987&0.320  &\textbf{6.567}& 4.381 \\
    PU-EVA~\cite{LuoPUEVAAE}  &0.571&5.840&3.937 &0.342&8.140&4.473 \\
    \hline  \hline
    Ours & \textbf{0.520} & 6.102 &\textbf{3.165}&\textbf{0.284} & 7.143 &\textbf{3.724} \\
    \hline
    \end{tabular}}
    \end{center}
\end{table}

\begin{figure*}[t]
    \centering
    \includegraphics[scale=0.28]{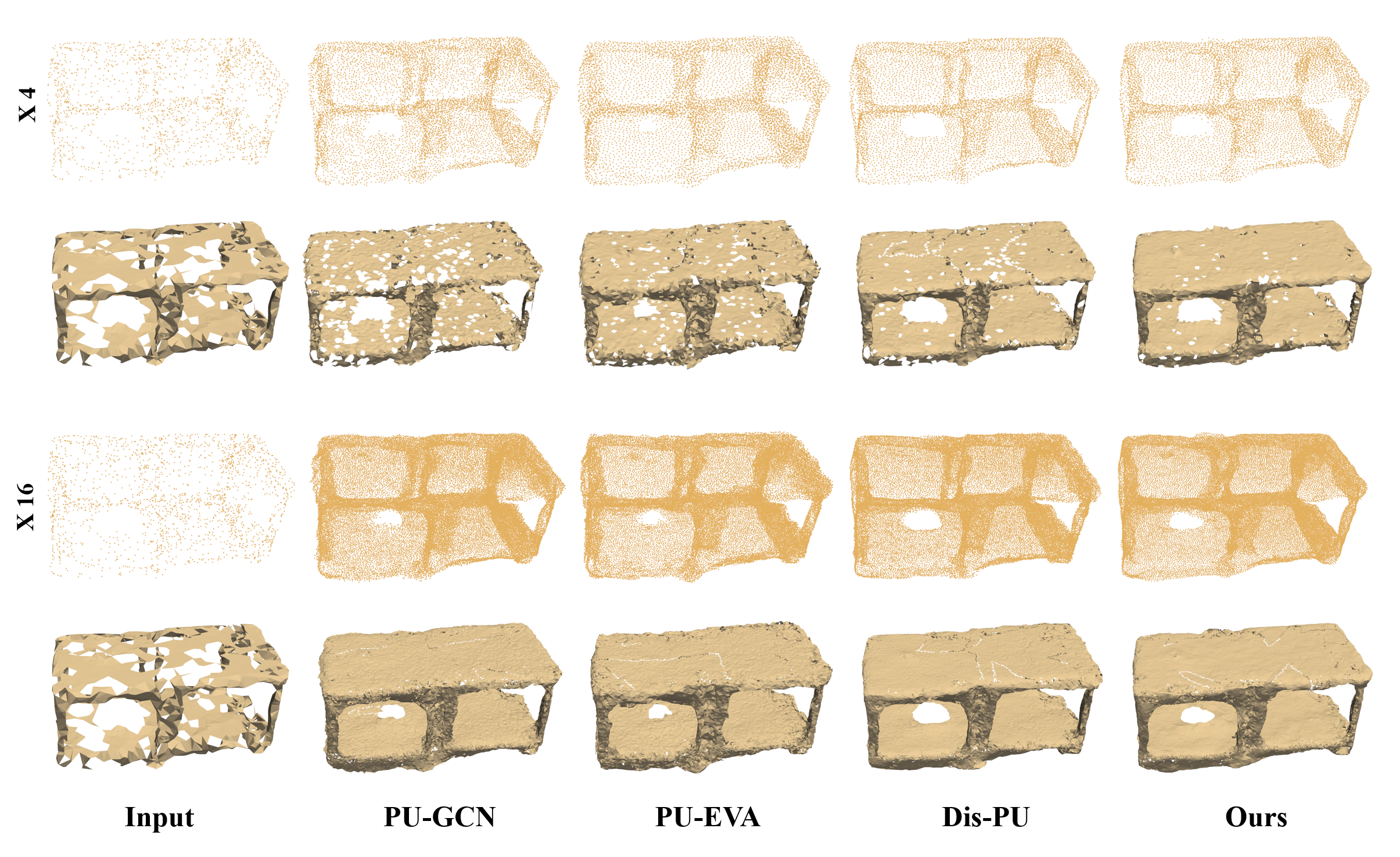}
    \caption{Point cloud upsampling ($\times 4$ and $\times 16$) results on real-scanned sparse inputs. Compared with the other counterparts, our method can generate more uniform with detailed structures. The 3D meshes are reconstructed by ball-pivoting algorithm~\cite{bernardini1999ball}.  }
    \label{realscanned_vis}
\end{figure*}

\noindent \textbf{PU-GAN dataset.} 
Following the previous settings~\cite{Li2021PointCU}, we also conduct experiments under two different upsampling rates at the input size of 1,024 on PU-GAN dataset.
Table~\ref{PU_GAN_performance} reports the quantitative results based on $\times$4 and $\times$16 upsampling rates. Since PU-GAN dataset only provides training patches based on $\times$4 upsampling rate, we apply the model twice for accomplishing $\times$16 upsampling rates in this experiment.
Overall, our method can achieve the best performance in most metrics. 
For Hausdorff Distance metric, some counterparts slightly outperform our method. We consider the most possible reason is that our transformer-based feature extraction module require more variations during the training. 
Compared with the 1,020 3D meshes in PU1K, PU-GAN is a much smaller dataset that only contains 120 meshes for training, which limits the capacity of the proposed method. 
Nonetheless, as for the results of Chamfer Distance and Point-to-Surface Distance, our method gains evident improvement over other counterparts.

\subsection{Results on Real-scanned Dataset}
\label{results_real}

To verify the effectiveness of our method in real-world scenarios, we utilize the models trained on PU1K and conduct the qualitative experiments on ScanObjectNN~\cite{Uy2019RevisitingPC} dataset. We only visualize the input sparse points and upsampled results, since ScanObjectNN dataset does not provide the corresponding dense points.
For clarity, we only choose three recent advanced methods (Dis-PU~\cite{Li2021PointCU},  PU-GCN~\cite{Qian2021PUGCNPC}, and PU-EVA~\cite{LuoPUEVAAE}) for visualization. 
As shown in Fig.~\ref{realscanned_vis}, we can observe upsampling the real-scanned data is more challenging, since the input sparse points are non-uniform and incomplete. 
Comparing with other counterparts, our method can generate more uniform and high-fidelity point clouds. The visualization of 3D meshes also shows the proposed is able to obtain a better surface reconstruction result. 
More visualization results on real-scanned data can be found in the supplementary material.

\subsection{Model Complexity Analysis}
\label{results_model}

In this section, we investigate the model complexity of our method compared with existing point cloud upsampling methods. The experiments are conducted on a Titan X Pascal GPU. 
As shown in Table~\ref{Complexity}, we can find that the proposed method is comparable with other counterparts. 
In point of training speed, our method is more efficiency than PU-GAN~\cite{Li2019PUGANAP}, Dis-PU~\cite{Li2021PointCU}, and PU-EVA~\cite{LuoPUEVAAE}. 
Note that Dis-PU~\cite{Li2021PointCU} employs a warm-up strategy during the training and requires more epoch to ensure the convergence of networks. Thus, the total training time of Dis-PU costs about 48h on PU-GAN dataset, which is much longer than ours (8h). 
As for inference time, we obtain a comparable and affordable inference speed compared with existing methods. 
In terms of model parameters, the proposed method can also be regarded as a lightweight network for point cloud upsampling.  
In addition, our method is scalable to achieve a trade-off between model effectiveness and efficiency by removing the refinement stage during the inference. In this manner, we can achieve leading performance as well as the model efficiency among other counterparts (the first bottom line). 
Overall, our cascaded refinement network can not only obtain the obvious performance improvements but also achieve comparable model complexity with other counterparts.

\begin{table}[t]
\caption{Comparison of model complexity with existing point cloud upsampling methods. The metrics of CD, HD, and P2F are calculated on PU1K dataset.}
    \label{Complexity}
    \centering
    \resizebox{1\linewidth}{!}{
    \begin{tabular}{|c||p{1.2cm}<{\centering}p{1.2cm}<{\centering}p{1.2cm}<{\centering}|c|c|c|}
    \hline
    Methods & \makecell[c] {{CD}}
   & \makecell[c] {{HD}} 
   & \makecell[c] {{P2F}}& \makecell[c] {~Training Speed~ \\ (s/batch)} &
        \makecell[c] {~Inference Speed ~\\ (ms/sample)} & ~Params. (Kb) ~ \\
         \hline \hline
        PU-Net~\cite{Yu2018PUNetPC}&1.157 & 15.297 & 4.924 &\textbf{0.13}&9.9 &812 \\
        MPU~\cite{Wang2019PatchBasedP3}&0.861 & 11.799 &3.181&0.23 &8.8 &\textbf{76}\\
        PU-GAN~\cite{Li2019PUGANAP} &0.661 &9.238 & 2.892&0.39 &13.1 &542  \\
        PU-GCN~\cite{Qian2021PUGCNPC}&0.635 &9.152 & 2.590  &\textbf{0.13}&8.6 &\textbf{76} \\
        Dis-PU~\cite{Li2021PointCU}&0.731&9.505&2.719&0.58 &11.2  &1047 \\
        PU-EVA~\cite{LuoPUEVAAE}&0.649&8.870&  2.715&0.57&12.8 & 2869\\
         \hline\hline
        Ours (w/o refiner) &0.492 &7.337&1.984&0.37 &\textbf{7.3} &567 \\
        Ours (full model)&\textbf{0.471} &\textbf{7.123}&\textbf{1.925}&0.37 &10.8 & 847 \\
        \hline
            \end{tabular}}

\end{table}

\subsection{Ablation Study}
\label{results_study}

To verify the effect of major components in our approach, we quantitatively evaluate the contribution of them. In the following, we provide the detailed discussions respectively.

\textbf{Ablation of generation stages.} 
Our full cascaded refinement network consists of two upsampling stage and one refinement  stage. 
In order to analyze the effectiveness of each sub-network, we investigate the setting of the number of generation stages. 
As shown in Table~\ref{ablation_stages}, the ``two-stage'' refers to remove the refinement stage, and the ``four-stage'' denotes adding an extra upsampling stage.  All the models are trained from scratch. Thus, the ``two-stage'' is different with merely removing the refiner in inference phase. 
We can find that the performance consistently increases with the number of generation stages. 
Our ``two-stage'' still outperforms the existing SOTA methods with a faster inference speed. 
In the view of the balance between effectiveness and efficiency, we adopt two upsampling stage and a refinement stage in this work.

\textbf{Effect of feature extraction module.} Feature extraction module plays an important role in point cloud upsampling. As shown in  Table~\ref{ablation_feature}, we investigate the effect of different network architecture on feature extraction. Specifically, ``MLP-based'' denotes the same feature extraction unit in MPU~\cite{Wang2019PatchBasedP3}, and ``DenseGCN'' refers to replace the point transformer layer with a DenseGCN block in~\cite{Li2019DeepGCNsCG}.
From the results, we can find that the proposed transform-based feature extraction module outperforms other counterparts, indicating its effectiveness of the locally self-attention operator on shape context learning. 
Moreover, the comparison of DenseGCN and MLP-based networks proves that an effective feature extraction network can promote obvious improvements on point cloud upsampling.

\begin{table}[t]
    \begin{center}
    \caption{Ablation of generation stages. The values of CD, HD, and P2F are multiplied by $10^{3}$.}
    \label{ablation_stages}
    \resizebox{0.7\linewidth}{!}{
    \begin{tabular}{|l||ccc|c|c|}
    \hline
  \multirow{2}{*}{Model}
    & \multicolumn{3}{c|}{{Dense (2048) input} } &
        \multirow{2}{*}{\makecell[c]{~Inference Speed ~\\ (ms/sample)}} &\multirow{2}{*}{Params. (Kb)} \\
  & \makecell[c] {{~~CD~~}}
  & \makecell[c] { {~~HD~~}} 
  & \makecell[c] {{~~P2F~~}} &&
      \\
    \hline\hline
    two-stage  &0.513& 7.808 & 2.078 &\textbf{7.3} & \textbf{567} \\
    three-stage &0.471 & 7.123& 1.925 & 10.8&847 \\
    four-stage &\textbf{0.466} & \textbf{6.793}& \textbf{1.821}& 14.7& 1127\\
    \hline
    \end{tabular}}
    \end{center}
\end{table}

\begin{table}[t]
\begin{minipage}{.40\linewidth}
    \begin{center}
    \caption{Effect of feature extraction module. }
    \label{ablation_feature}
    \resizebox{0.9\linewidth}{!}{
    \begin{tabular}{|l||ccc|}
    \hline
  \multirow{2}{*}{Model}
    & \multicolumn{3}{c|}{{Dense (2,048) input}~ }  \\
  &  \makecell[c] {{~~CD~~}}
  & \makecell[c] { {~~HD~~}} 
  & \makecell[c] {{~~P2F~~}}
      \\
    \hline\hline
    MLP-based & 0.628 & 9.687 & 2.721  \\
    DenseGCN & 0.531 & 8.462 & 2.254  \\
    Ours &\textbf{0.471} & \textbf{7.123}& \textbf{1.925} \\
    \hline
    \end{tabular}}
    \end{center}
    \end{minipage}
\begin{minipage}{.59\linewidth}
    \begin{center}
    \caption{Effect of residual learning scheme. }
    \label{ablation_learning}
    \resizebox{0.9\linewidth}{!}{
    \begin{tabular}{|c|c|ccc|}
    \hline
  \multirow{2}{*}{~~Warm-up~~}
  & \multirow{2}{*}{~Residual learning~} 
  & \multicolumn{3}{c|}{{~Dense (2,048) input}~}  \\
  & &  \makecell[c] {{~~CD~~}}
  & \makecell[c] { {~~HD~~}} 
  & \makecell[c] {{~~P2F~~}}
      \\
    \hline \hline
    \checkmark& - &0.511& 7.599 &2.986 \\
    \checkmark & \checkmark &\textbf{0.472}& \textbf{7.329}&
    \textbf{1.933} \\   
    \hline \hline
    -& -&0.733 & 9.667& 3.380  \\
    -& \checkmark  &\textbf{0.471} & \textbf{7.123}& \textbf{1.925} \\
    \hline
    \end{tabular}}
    \end{center}
    \end{minipage}
\end{table}

\textbf{Study of residual learning.} 
For a coarse-to-fine upsampling framework, it is crucial to mitigate the learning conflicts between multiple generation units. 
As mentioned in Section~\ref{discussion}, MPU needs to progressively activate the training of each upsampling unit, and Dis-PU employs a warm-up strategy that gradually increases the learning rate of spatial refiner. 
In contrast, we only adopt the residual learning in each generation stage without progressive or warm-up training strategy. 
From the results in Table~\ref{ablation_learning}, we can find that residual learning is more effective than warm-up training strategy in our framework.
The reason behind is, that the residual connection makes each stage only regress the point offsets of the initial shape and thus mitigate the learning conflicts between multiple stages, resulting in a better optimization.  
Therefore, we equip the residual learning for point reconstruction, which makes the proposed method be easily optimized without extra training strategies.

\section{Conclusion}

In this paper, we propose a cascaded refinement network for point cloud upsampling. 
The proposed method consists of three generation stages, which progressively generates the coarse dense points and refine the coarse points to a better position in the end. 
We adopt a residual learning scheme for point coordinate reconstruction, which enables to decrease the difficulty for regressing the new points and mitigate the conflicts between multiple stages.
Moreover, a transformer-based feature extraction module is designed to aggregate the global and local shape context, and a two-branch based feature expansion module enables to learn a more expressive upsampled features. 
Compared with the existing coarse-to-fine point upsampling methods, the proposed network can be easily optimized without extra training strategies. 
Moreover, we can also dynamically adjust the model efficiency and effectiveness at the stage of inference, depending on the available computational resources. 
Both quantitative and qualitative results demonstrate the superiority of our method over the state-of-the-art methods.

\bibliographystyle{splncs04}
\bibliography{egbib}

\appendix
\section{Appendix}
In the following, we provide detailed network architecture and more experiments results to support the main paper. 

\section{Network Details}
In this section, we elaborate the detailed network architecture. 

\noindent \textbf{Feature extraction.} 
In Fig.3 of the main text, we provide the overall structure of feature extraction module. Here, we give more details about the parameter settings. 
At the start of the feature extraction module, we feed the initial points to a set of MLPs, in which the numbers of output channel are 64, and 128, respectively. 
Then, the global feature is produced through a max pooling operation. After that, the duplicated global features are concatenated with the initial features and reduced to 128 channels via two MLP layers. 
We further adopt a point transformer layer~\cite{Zhao2020PointT} to refine the local shape context, and the channel number $C'$ is set to 64 at the stage of feature transformations. 
After the local self-attention operation, the channel number of output point features is 128. 

\noindent \textbf{Feature expansion.} 
As presented in the main text, we employ two branches for feature expansion. For  transposed convolution-based  branch, we set the output channel numbers of MLPs as 32, and then a one-dimensional deconvolution layer is utilized to produce expanded features with 128 channel numbers and $r$ times point numbers. 
For duplicate-based branch, the output features also have $r$ times point numbers with 128 output channel numbers.
Then, we concatenate the two-branch features and obtain the expanded features using two MLP layers, in which the output channel numbers are 256 and 128, respectively. 

\noindent \textbf{Coordinate reconstruction.} For regressing the per-point offset $\Delta \mathcal{P}$, the expanded features are gradually reduced to 64 and 3 channels through two MLP layers. Then, the per-point offset $\Delta \mathcal{P}$ is added on the $r$ times duplicated input point clouds.

\section{More Experimental Results}
In this section, we present more experimental results, including effect of training supervision, ablation study on refinement stage and visualization results on real-scanned data.

\subsection{Effect of Training Supervision}
In this section, we conduct an experiment to verify the effectiveness of the supervision on three stages. 

\begin{table}[t]
    \begin{center}
    \caption{Effect of training Supervision. The values of CD, HD, and P2F are multiplied by $10^{3}$. A smaller value denotes a better performance. }
    \label{ablation_Supervision}
    \resizebox{0.7\linewidth}{!}{
    \begin{tabular}{|l||p{1cm}<{\centering}p{1cm}<{\centering}p{1cm}<{\centering}|p{1cm}<{\centering}p{1cm}<{\centering}p{1cm}<{\centering}|}   
        \hline

   \multirow{2}{*}{Model}
   & \multicolumn{3}{c|}{{Medium (1,024) input}} 
   & \multicolumn{3}{c|}{{Dense (2,048) input} }  \\ 
   & \makecell[c] {{CD}}
   & \makecell[c] {{HD}} 
   & \makecell[c] {{P2F}}
   & \makecell[c] {{CD}}
   & \makecell[c] {{HD}} 
   & \makecell[c] {{P2F}}
   \\ \hline\hline
    Last stage&0.832&11.443&3.108&0.507 & 7.904 & 1.972 \\
   \hline  
    All stages &\textbf{0.808}&\textbf{10.750}&\textbf{3.061}& \textbf{0.471} &\textbf{7.123}&\textbf{1.925} \\
    \hline
    \end{tabular}}
    \end{center}
\end{table}

From the results in Table~\ref{ablation_Supervision}, we can see that adding the supervision on all stages achieves a better result than only constraining on the last stage. 
We consider the reason behind is that the supervision on each stage enables to make its output more reliable and then provides a better initial shape for the next stage. 
Therefore, we calculate CD loss for three generation stages and optimize them simultaneously.

\subsection{Ablation Study on Refinement Stage}

\begin{figure}[ht!]
    \centering
    \includegraphics[scale=0.21]{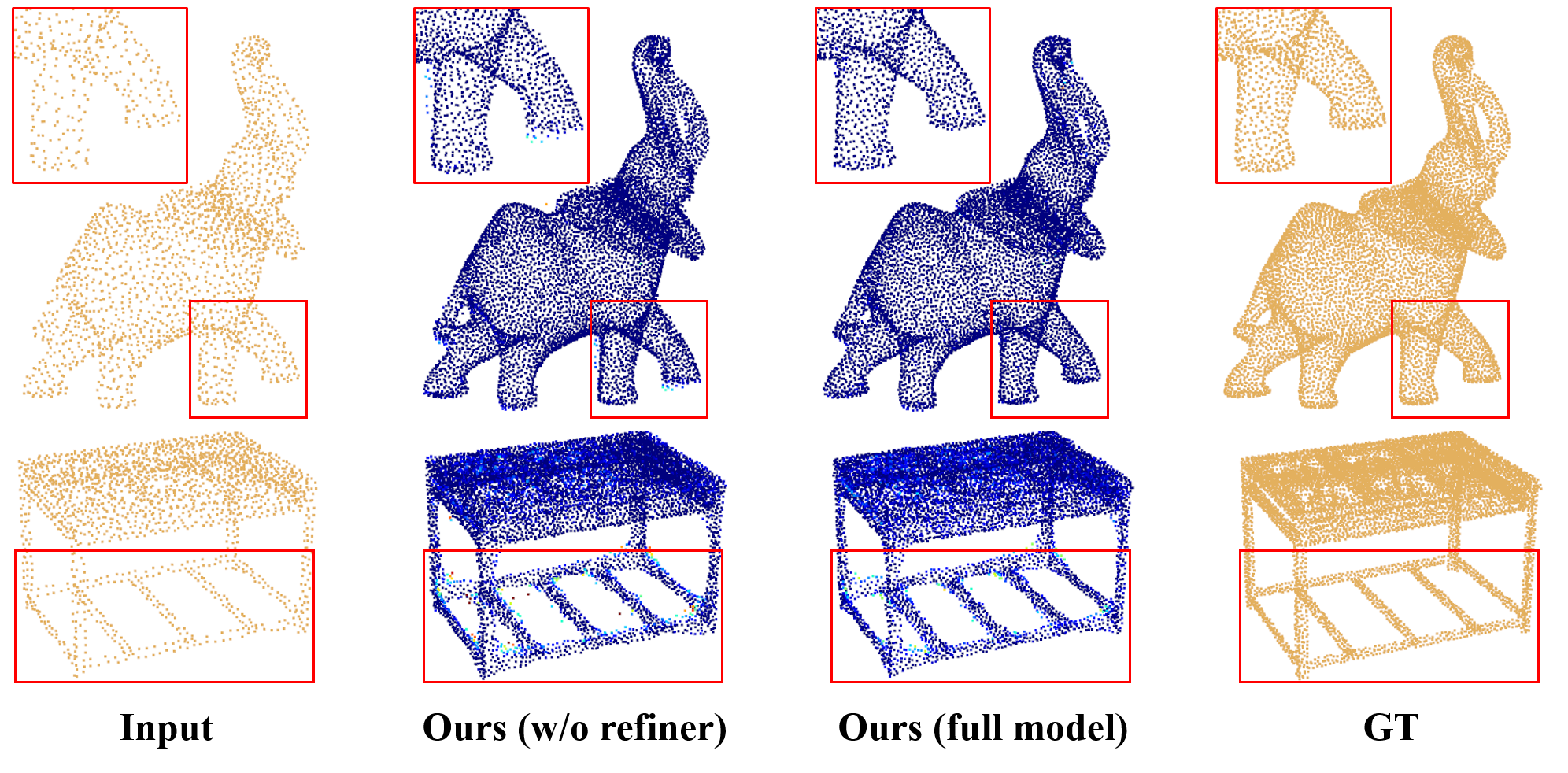}
    \caption{Qualitative comparisons on refinement stage. }
    \label{ablation}
\end{figure}

Fig.~\ref{ablation} provides some visualized results on removing the refinement stage in the inference phase. From the results, we can find there are some outliers produced by the second upsampling stage (the second column). Then, the refiner enables to adjust them to a better position, and thus obtains a result with higher fidelity.

\begin{figure}[t]
    \centering
    \includegraphics[scale=0.23]{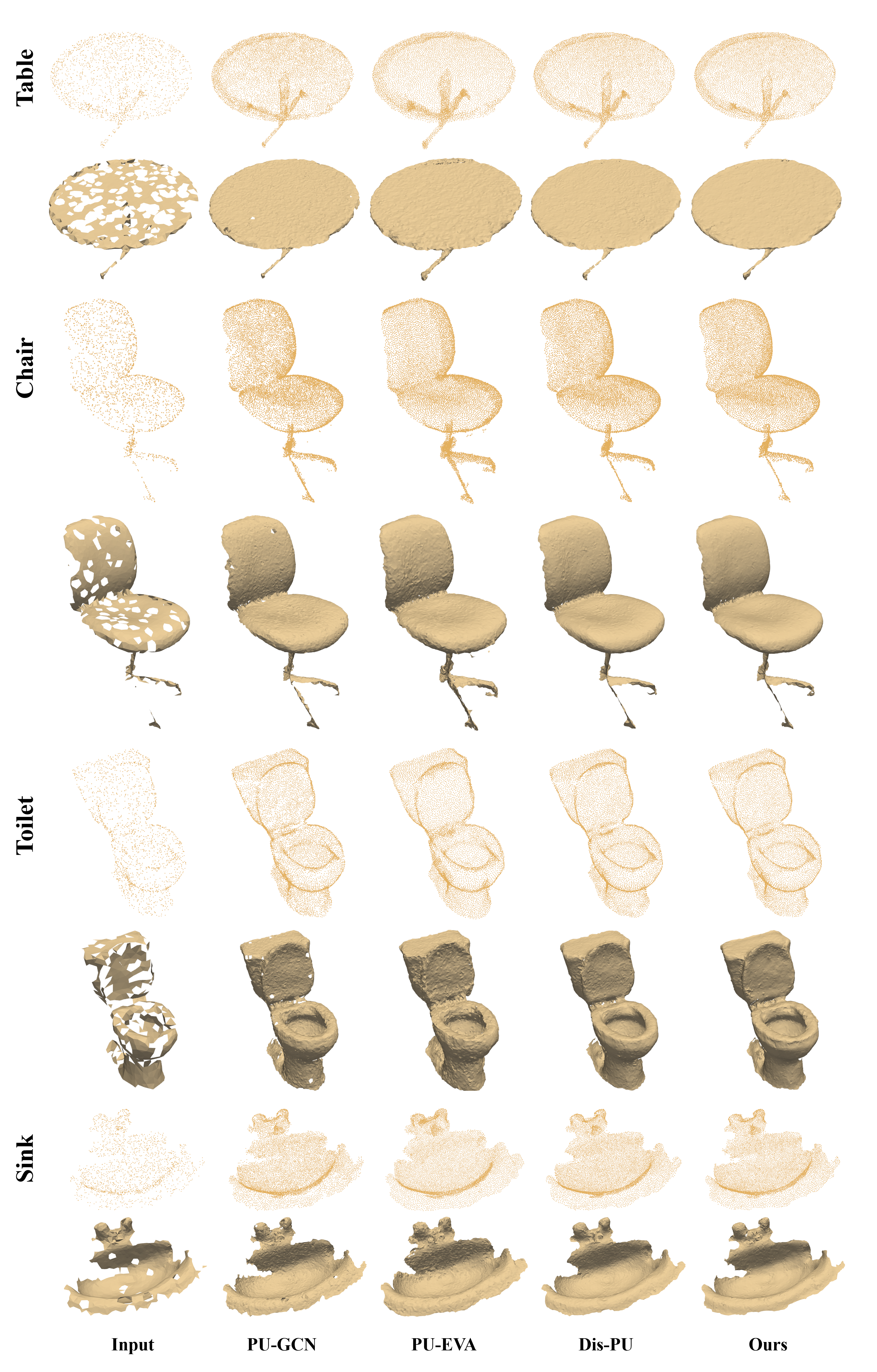}
    \caption{Point cloud upsampling ($\times 4$) results on real-scanned sparse inputs. Compared with the other methods, our upsampled point clouds are more uniform and proximity-to-surface. One can zoom in for details.}
    \label{realscanned_vis_2}
\end{figure}

\subsection{Visualization Results}
Here, we give more visualization results on ScanObjectNN~\cite{Uy2019RevisitingPC} dataset. 
As shown in Fig.~\ref{realscanned_vis_2}, our method  generate more uniform with detailed structures on various objects compared with other competitors. 
The 3D surface reconstruction are largely influenced by the quality of the upsampled point clouds. The proposed method is able to preserve the details in the sharp areas and smoothness in the smooth regions. The visualization results demonstrate that our upsampled point clouds are more uniform and close to the target surface.

\end{document}